\documentclass[A4paper, 11 pt, conference]{ieeeconf}  
\IEEEoverridecommandlockouts       
\usepackage{geometry}
\usepackage{graphics} 
\usepackage{epsfig} 
\usepackage{mathptmx} 
\usepackage{times} 
\usepackage{amsmath} 
\usepackage{amssymb}  
\usepackage{subfigure}
\usepackage{algorithm} 
\usepackage{algorithmic} 
\usepackage{hyperref}

\usepackage{multirow} 
 
\begin{document}

\title{\LARGE \bf
Collaborative Fall Detection and Response using Wi-Fi Sensing and Mobile Companion Robot}

\author{Yunwang Chen$^{\dag,1}$, Yaozhong Kang$^{\dag,2}$, Ziqi Zhao$^{1}$, Yue Hong$^{1}$, Lingxiao Meng$^{1}$, and Max Q.-H. Meng$^{*, 1}$%
\thanks{All authors are with the Shenzhen Key Laboratory of Robotics Perception and Intelligence, Southern University of Science and Technology, Shenzhen 518055, China.}
\thanks{${}^1$Yunwang Chen, Ziqi Zhao, Yue Hong, Lingxiao Meng and Max Q.-H. Meng are also with the Department of Electrical and Electronic Engineering, SUSTech, Shenzhen, China. Emails: \{chenyw2021, zhaozq2020, 12331232, menglx2021\}@mail.sustech.edu.cn and max.meng@ieee.org.}
\thanks{${}^2$Yaozhong Kang is also with the School of System Design and Intelligent Manufacturing, SUSTech, Shenzhen, China. Email: kangyz2021@mail.sustech.edu.cn.}
\thanks{${}^*$Corresponding author: Max Q.-H. Meng.}
\thanks{$^{\dag}$The first two authors contributed equally to this work.}}

\maketitle 
\thispagestyle{empty}

\begin{abstract}
This paper presents a collaborative fall detection and response system integrating Wi-Fi sensing with robotic assistance. The proposed system leverages channel state information (CSI) disruptions caused by movements to detect falls in non-line-of-sight (NLOS) scenarios, offering non-intrusive monitoring. Besides, a companion robot is utilized to provide assistance capabilities to navigate and respond to incidents autonomously, improving efficiency in providing assistance in various environments. The experimental results demonstrate the effectiveness of the proposed system in detecting falls and responding effectively.
\end{abstract}


\section{Introduction}

As people age, they often experience various issues such as mobility decline, cognitive impairment, and physical health deterioration. For the elderly, falls are particularly detrimental, often resulting in long-term health complications and a diminished quality of life. Indoor falls are especially problematic due to the potential lack of immediate assistance, leading to prolonged periods before help arrives, exacerbating injuries and complicating recovery. Consequently, they must rely on the assistance of family members and caregivers, which creates a significant burden on their families \cite{fall2024}. 

Providing efficient, cost-effective non-line-of-sight (NLOS) home healthcare for this growing group of older adults has a profound societal impact, in which accurate and prompt indoor fall detection is crucial \cite{FALLSOUND}. For caregivers, it is essential to know if an elder has fallen behind an obstacle, such as a closed door . Traditional methods primarily utilize wearable devices equipped with accelerometers and gyroscopes to monitor sudden changes in motion and orientation. These sensors effectively detect rapid movements indicative of falls, providing real-time alerts to caregivers or emergency services \cite{FallWear}. However, these devices often face user compliance issues; elderly individuals may forget to wear them or find them uncomfortable, leading to inconsistent usage and unreliable monitoring. Vision-based systems, which need to be installed indoors, employ cameras and image processing to detect falls by analyzing visual data \cite{VisionFall2024}. Despite their accuracy, these systems raise  privacy concerns due to constant surveillance in private spaces, posing a critical barrier to their widespread adoption .
Recent advancements in Wi-Fi sensing have demonstrated potential for human activity recognition by analyzing disruptions in Wi-Fi signals caused by movements \cite{yousefi2017survey}. Wi-Fi sensing allows widely used commercial Wi-Fi access points (APs) installed outside obstacles to detect falls inside using the channel state information (CSI) between the Wi-Fi AP and other Wi-Fi devices, effectively addressing the need for NLOS detection. This non-intrusive approach can detect falls without compromising privacy, making it a promising solution for sensitive areas such as bathrooms. 

To quickly reach the patient and provide emergency treatment, mobile manipulator systems have emerged as a promising solution for providing timely assistance to the elderly \cite{bardaro2022robots}. Typical mobile manipulators are equipped with navigation and manipulation capabilities, enabling them to autonomously perform a variety of tasks that support independent living. By integrating these robotic advancements with Wi-Fi sensing technology, we can  improve the efficiency and quality of emergency responses.

In this paper, we propose a novel fall detection and response system that integrates Wi-Fi sensing with a patrolling robot capable of providing assistance. The Wi-Fi sensing device detects falls by analyzing CSI disruptions in Wi-Fi signals caused by human movements. Specifically, we utilize the amplitude of Wi-Fi CSI provided by commodity Wi-Fi devices along with deep learning models to detect falls. The patrolling robot can autonomously respond to fall incidents, navigate indoor environments including door traversal and offer assistance to the fallen person. The test experiment shows that this integrated approach is capable of providing continuous, privacy-preserving monitoring and immediate, autonomous assistance.


The rest of the paper is organized as follows: Section II reviews related work on Wi-Fi sensing and mobile companion robots. Section III describes the methods and system design, including the Wi-Fi sensing module and the companion robot. Section IV details the real-world experiments conducted to evaluate the system, including the experimental setup and results. Finally, Section V concludes the paper and discusses future work.

\section{Related Work}
\subsection{Wi-Fi Sensing using CSI}

Wi-Fi devices based on the IEEE 802.11a/g/n/ac standards employ orthogonal frequency division multiplexing (OFDM) as the modulation scheme, featuring multiple sub-carriers in a Wi-Fi channel and multiple antennas to mitigate frequency-selective fading. The receiver measures a discrete channel frequency response (CFR) over time and frequency as phase and amplitude, encapsulated in the form of CSI for each antenna pair.
 In wireless communication, CSI is know as the channel property of a wireless communication channel. CSI captures the effects of various phenomena, including reflection, scattering, and fading, which occur as a Wi-Fi signal propagates through an environment. By measuring the amplitude and phase of the signal at different subcarriers between the transmitter and the receiver, CSI provides a comprehensive description of how the signal is modified. This information can be utilized to infer the presence and movements of objects within the environment \cite{yousefi2017survey}. Mathematically, the relationship between the transmitted signal \( \mathbf{X} \) and the received signal \( \mathbf{Y} \) is represented as:
\begin{equation}
\mathbf{Y} = \mathbf{H}\mathbf{X} + \mathbf{N}
\end{equation}
where \( \mathbf{Y} \in \mathbb{C}^{N_r \times \delta t} \) is the received signal matrix, \( \mathbf{H} \in \mathbb{C}^{N_r \times N_t} \) is the CSI matrix representing the channel effects, \( \mathbf{X} \in \mathbb{C}^{N_t \times \delta t} \) is the transmitted signal matrix, and \( \mathbf{N} \in \mathbb{C}^{N_r \times \delta t} \) denotes the noise matrix. Here, \( N_r \) represents the number of receiving antennas, \( N_t \) represents the number of transmitting antennas, and \( \delta t \) is the length of the communication frame.

Recent advancements in Wi-Fi sensing have employed deep learning techniques to process CSI data for human activity recognition (HAR). El Zein et al. achieved over 90\% accuracy using deep CNNs and time series data augmentation for real-time HAR \cite{ElZein2023}. Su et al. employed multilayer bidirectional LSTM networks with self-powered sensors, achieving over 96\% accuracy \cite{Su2022}. Mekruksavanich and Jitpattanakul's CSI-ResNeXt network achieved 99.17\% accuracy with lightweight deep residual networks \cite{Mekruksavanich2023}. However, most of these works lack real-world experiments for testing and validation datasets and also lack deployment in real-world NLOS scenarios.

\subsection{Mobile Manipulator System}

Several studies have highlighted the effectiveness of mobile service robots equipped with navigation and manipulation capabilities in providing timely assistance to the elderly \cite{bardaro2022robots}.
For single tasks such as door traversal, recent works \cite{8793866} \cite{10132587} \cite{arduengo2021robust} have proposed solutions using either pre-planned methods or sensing-based approaches to handle obstacles in the way. For long-horizon tasks, the Mobile ALOHA project has introduced a low-cost mobile manipulation system capable of performing complex, through whole-body teleoperation and imitation learning \cite{fu2024mobile}. Furthermore, the multi-skill mobile manipulation approach proposed by \cite{gu2022multi}, integrates mobility with manipulation skills and introduces a region-goal navigation reward, demonstrating superior performance in long-horizon mobile manipulation tasks.

These studies highlight the potential for effectively managing tasks related to elderly care. By integrating Wi-Fi sensing as an alert, a mobile companion robot system can improve emergency response efficiency and overall care quality for older adults.


\section{Methods}
\begin{figure*}[t]
    \centering
    \includegraphics[width=\textwidth]{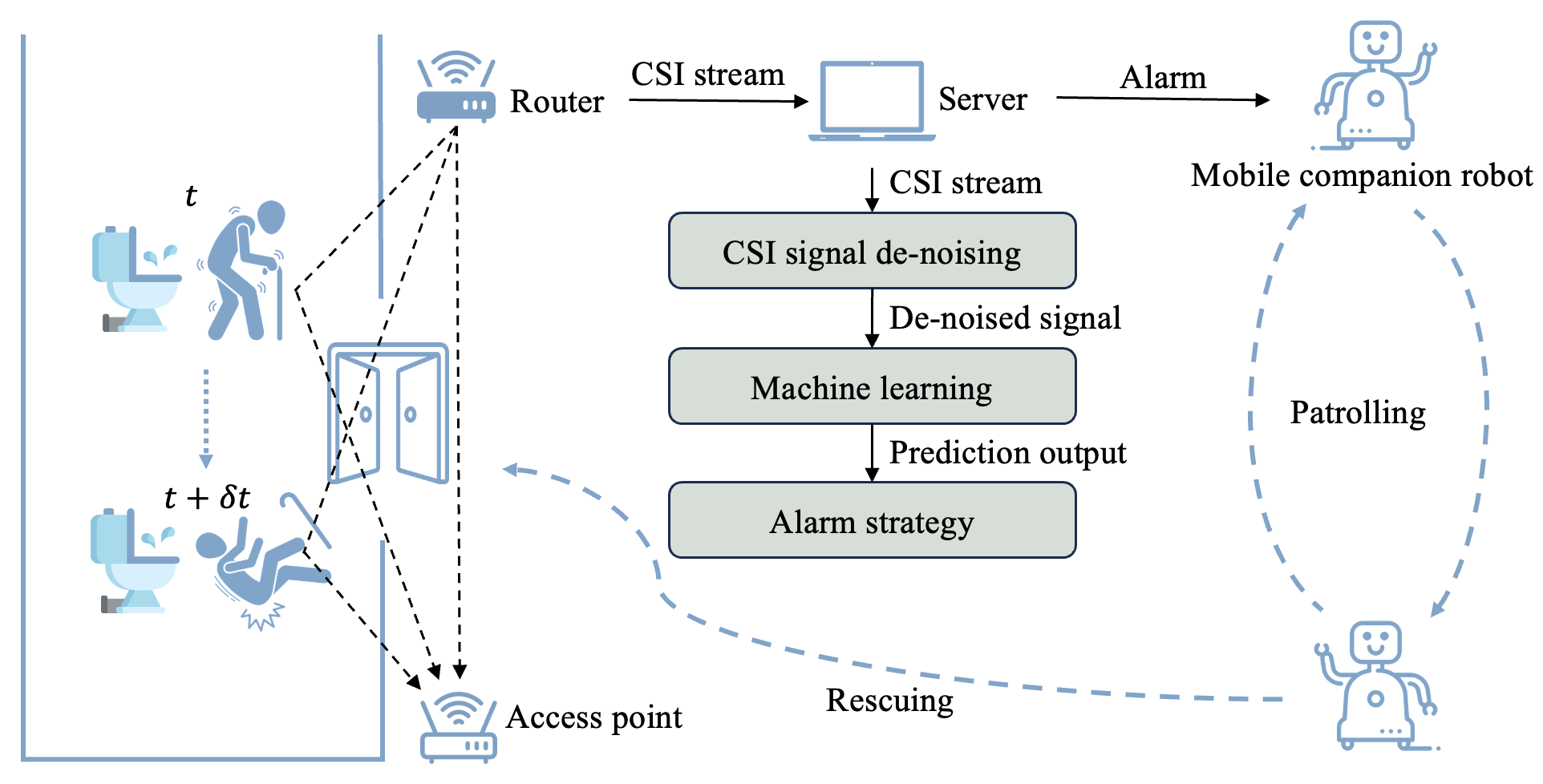}
    \caption{System Overview}
    \label{fig:SystemOverview}
\end{figure*}

The proposed system consists of two main components: the Wi-Fi sensing module and the mobile companion robot, as shown in Fig. 1. The Wi-Fi sensing module detects falls by analyzing signal disruptions caused by movements. The mobile companion robot, equipped with navigation and manipulation capabilities, responds to detected falls by providing assistance and contacting caregivers if necessary.

\subsection{Wi-Fi Sensing Component}

Wi-Fi sensing leverages CSI to identify anomalies in Wi-Fi signals, which can indicate events such as falls. We adopt the two-stream convolution augmented transformer model as our base model, as proposed by \cite{bing2021that}. This model captures intricate patterns and dependencies in the data that traditional machine learning methods may overlook, in which LSTM networks are adept at modeling temporal sequences in CSI data, while CNNs excel in identifying spatial features across CSI measurements. The base model integrates both CNN and LSTM with transformer multihead strategy, significantly enhancing the accuracy and robustness of activity recognition systems, making them more feasible for real-world applications.

During the training process, we collected real CSI data corresponding to three distinct states: 'Fall', 'Normal', and 'No-person/Static' in our NLOS experiment setup, as detailed in section \ref{Setup}. To enhance the robustness and accuracy of our model, we employed transfer learning techniques. Specifically, we pre-trained the transformer model using the dataset from \cite{yousefi2017survey} and then fine-tuned it by modifying the original final layer with new data collected, as detailed in section \ref{transferlearning}.

\subsection{Mobile Companion Robot}
\subsubsection{Hardware Design}
\begin{figure}[htbp]
    \centering
    \includegraphics[width=0.7\columnwidth]{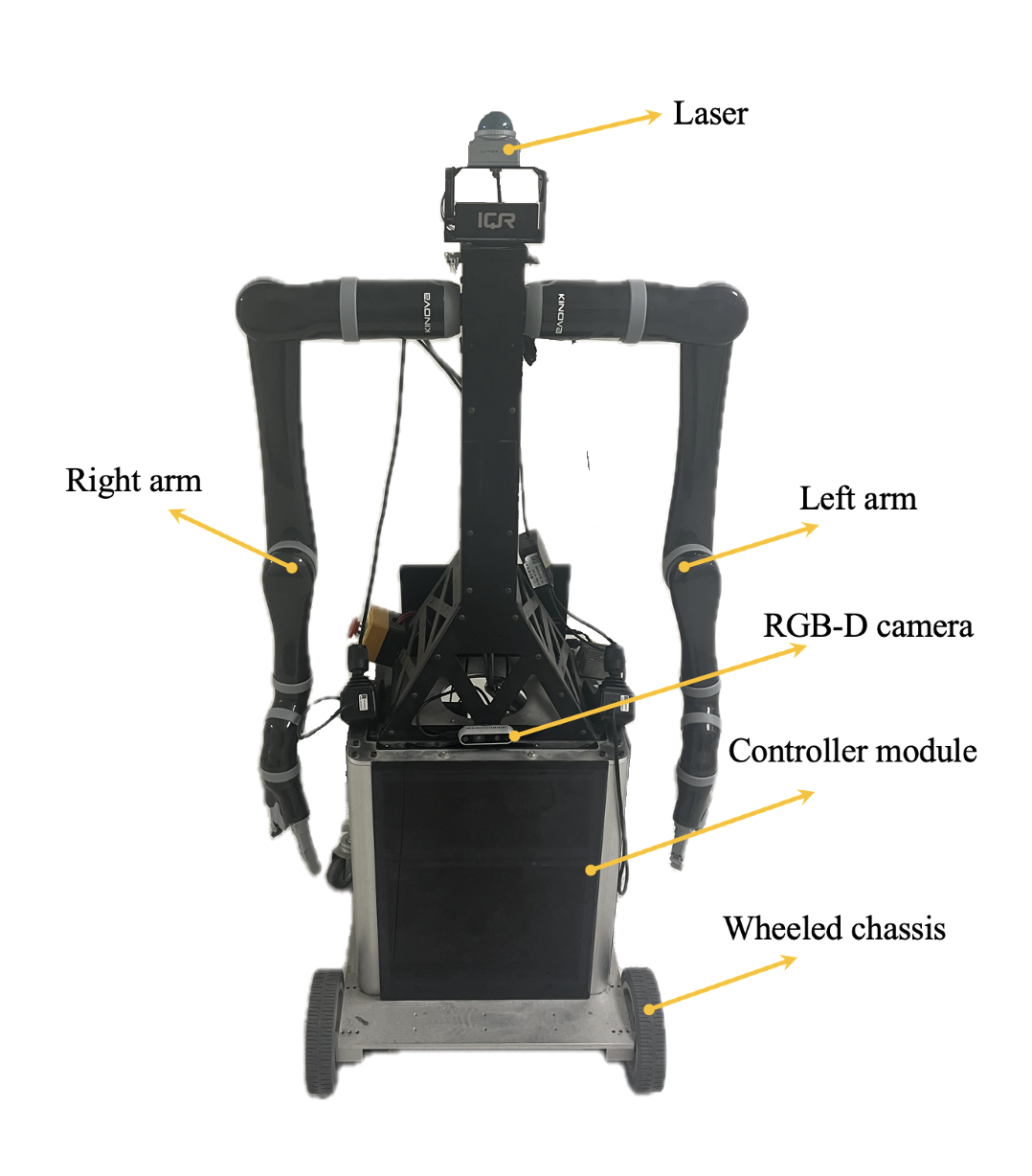}
    \caption{Overview of the dual-arm mobile companion robot}
    \label{fig:mobile_companion_robot}
\end{figure}

Our mobile robot features a differential drive base with two power wheels, dual robotic arms, an RGB-D camera, and a laser \ref{fig:mobile_companion_robot}. The dual-arm mobile features two 7-DOF simulation arms, giving it ability to handle complicated missions such as freeing itself from being obstructed or performing first aid tasks. The total reachable workspace and the shared workspace of the dual arms are identified in \cite{9517634}.

\subsubsection{Software Design}
\begin{figure*}[t]
    \centering
    \includegraphics[width=0.8\textwidth]{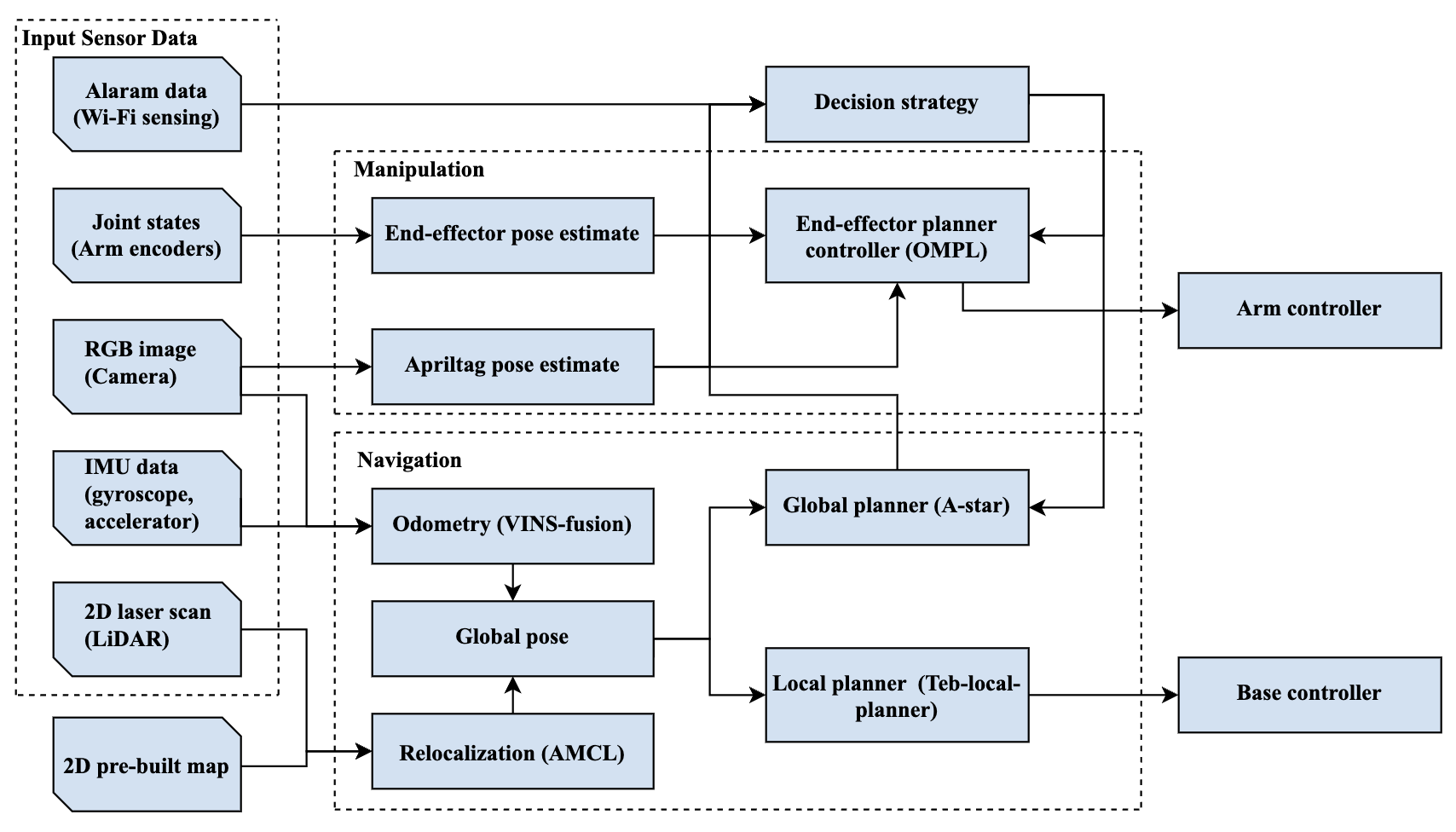}
    \caption{Manipulation and navigation system design overview}
    \label{fig:mani_navi}
\end{figure*}

The robot utilizes various sensors to collect data for its operation. Alarm data, containing the status of the subject elderly, is obtained through Wi-Fi sensing. Joint states are obtained from arm encoders to provide precise absolute angles of the joints in order to obtain the pose of the end-effecter through forward kinematics. The RGB-D camera captures image data for object recognition and provide video information for remote assessing. IMU (inertial measurement unit) data including gyroscope and accelerometer readings, obtained from the RGB-D camera's IMU module, aids in the pose estimation of the body. Additionally, the laser scanner (LiDAR) offers detailed distance measurements to create a map of the surroundings using gmapping, and relocalize itself in a pre-built map using AMCL \cite{4084563}.In addition, the control system is divided into two main controllers: the arm controller and the base controller. The arm controller executes the planned trajectories for the robotic arms, while the base controller manages the navigation and movement of the robot chassis.

The data collected from these sensors undergoes several processing steps to ensure accurate localization and planning. AprilTag pose estimation is performed on the RGB image data to enhance visual localization of the precise 3D pose of the door and therefore localize the door handle. The odometry of the robot's body is calculated using VINS-Fusion \cite{qin2017vins}, which combines visual and inertial data to provide accurate state estimations under scenarios that will degrade common lidar-wheeled fusion estimate algorithms.

For inverse kinematics and arm control, the system integrates multiple planning modules. The end-effector planner and controller utilize OMPL \cite{moll2015benchmarking-motion-planning-algorithms} to generate feasible trajectories for the robotic arms. Global path planning for chassis is handled by the A-star algorithm, ensuring an optimal route through the environment. Under the differential wheel structure, the teb-local-planner exhibits excellent navigation performance \cite{Rsmann12}, which dynamically adjusts the path based on real-time obstacles and robot constraints.

\section{Experiments}

\subsection{Experimental Setup}
\label{Setup}
\begin{figure}[htbp]
    \centering
    \includegraphics[width=0.8\columnwidth]{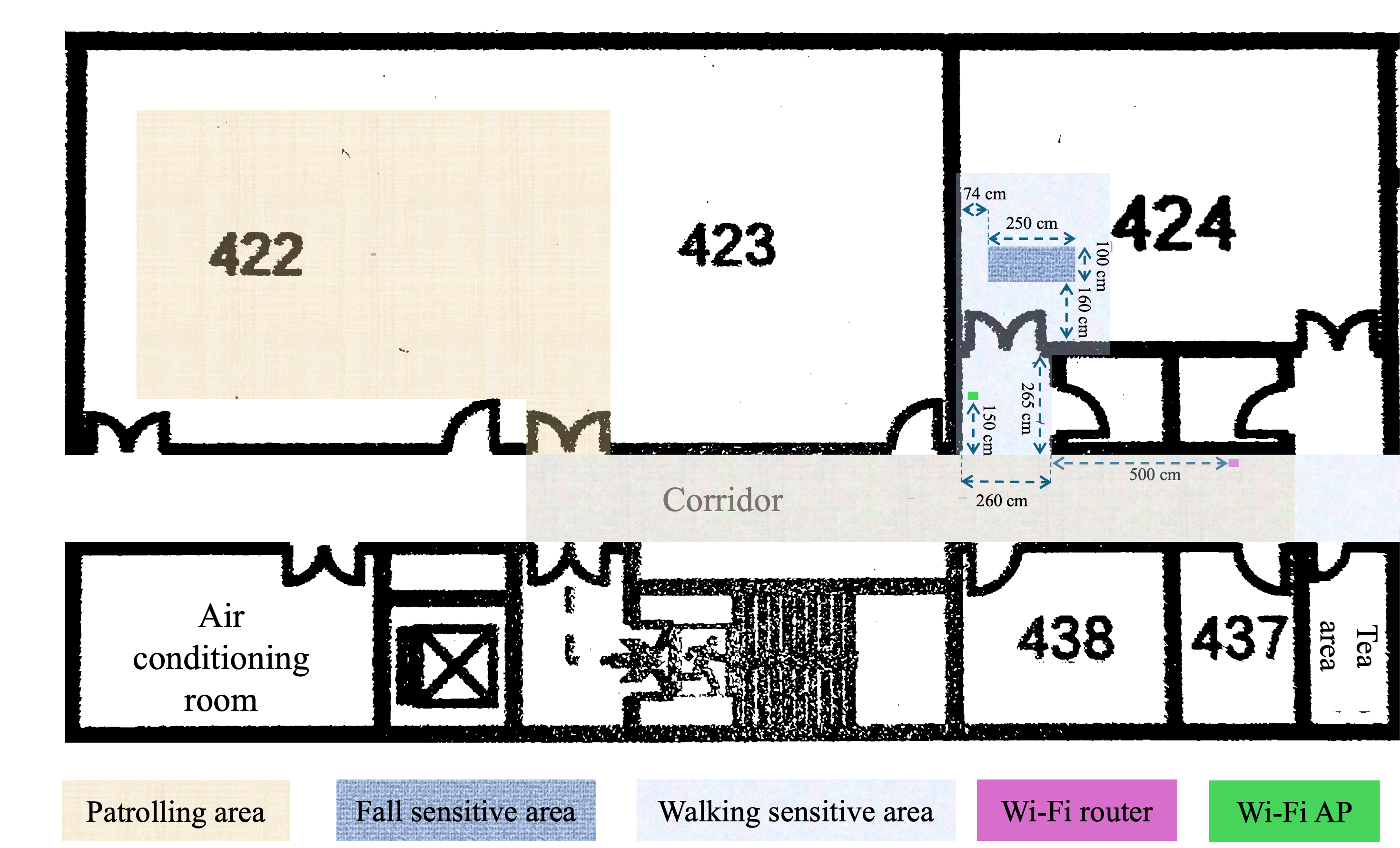}
    \caption{Experiment setup}
    \label{fig:floor}

\end{figure}

\begin{figure}[htbp]
    \centering
    \includegraphics[width=0.8\columnwidth]{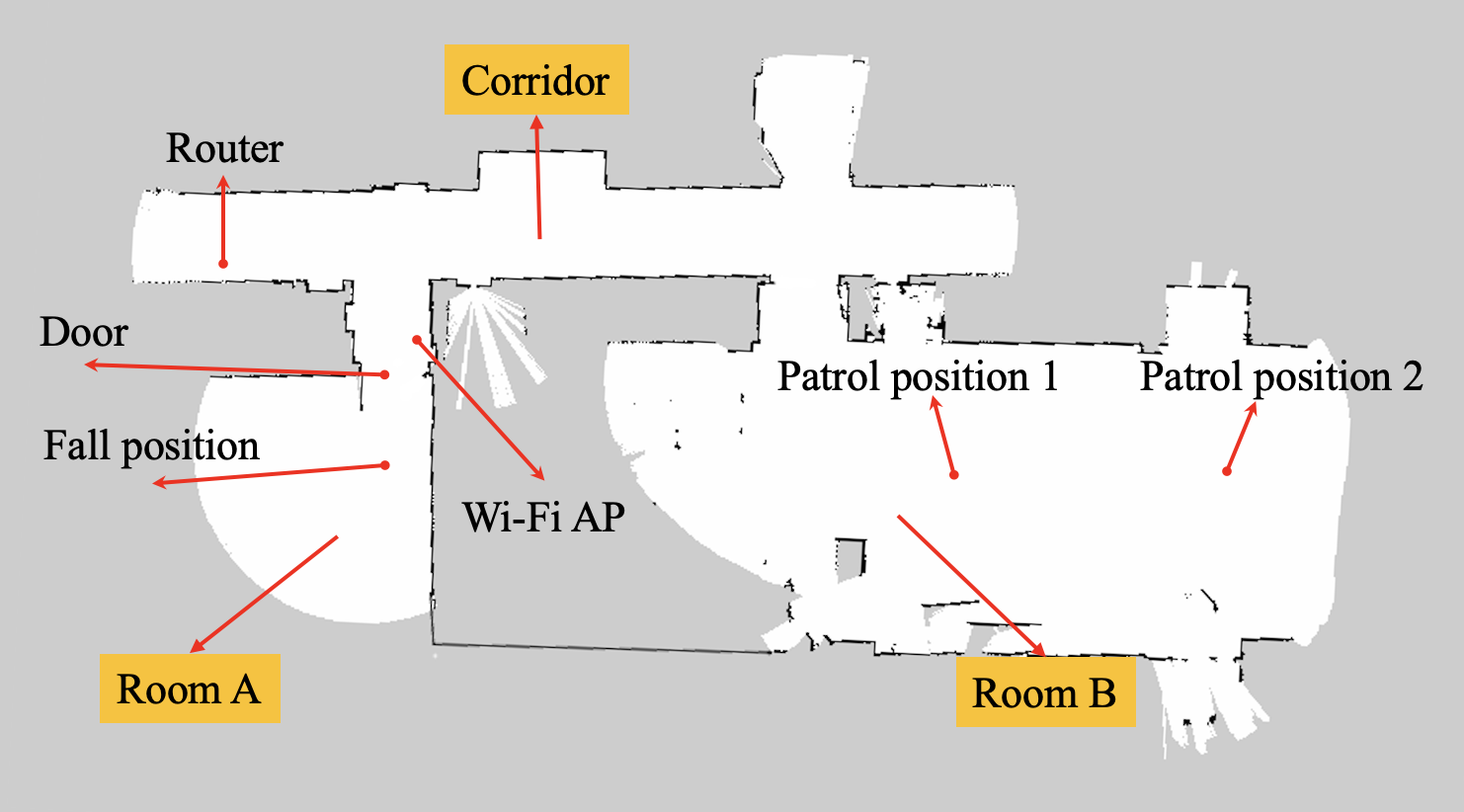}
    \caption{Experiment map}
    \label{fig:exper_map}
\end{figure}

Accessing CSI is not supported by all commercially available IEEE 802.11 chipsets. To overcome this limitation, we utilize the Intel 5300 network interface card (NIC), which is well-regarded for its ability to extract CSI from the physical layer with the assistance of the Linux 802.11n CSI Tool \cite{Halperin_csitool}. Our setup includes a TP-Link AX6000 (TL-XDR6020) Wi-Fi router and a Lenovo AX 201 laptop equipped with an Intel 5300 NIC for CSI data retrieval. The laptop can be regarded as  a Wi-Fi AP or a Wi-Fi amplifier. Both the router and the laptop  are positioned outside a conference room as illustrated in Fig. \ref{fig:floor}.
Fig. \ref{fig:floor} also shows the experiment setup, including various designated areas such as the patrolling area, fall sensitive area, walking sensitive area.

Additionally, a Macbook with an M1 Pro chip is used as a server to process Wi-Fi CSI data and a companion robot is patrolling waiting for fall signal. The robot uses a VLP-16 as laser input, Intel \textit{Realsense} D435-i as RGB image and IMU data input, and two Kinova Gen2 robotic arms, as shown in Fig. \ref{fig:mobile_companion_robot}.  Furthermore, Fig. \ref{fig:exper_map} presents the floor map generated by SLAM.

\subsection{Experiment and Results}
\label{transferlearning}

In the base model training, we used the dataset from \cite{yousefi2017survey}, where the Wi-Fi transmitter and receiver were located 3 meters apart in a line-of-sight (LOS) condition and included 7 categories of action (total of 557 sets), as shown in Table \ref{originalDataset}. Although this dataset is not perfectly aligned with our environment setup, we used it to train the base model because it helps the model learn the features of each action and eases further training. The base dataset was divided into a training dataset and a test dataset with a ratio of 8:2, and fed into the base model. We saved the model at epoch 34, as it attained the highest test accuracy of 90.1\%, and used it as the pretrained model for transfer learning.

To adapt the model to our specific objectives and given our limitations in collecting the dataset, we collected a total of 135 sets of experimental data with 3 categories, as shown in Table \ref{transferLearningDataset}, with walking and fall areas detailed in Fig. \ref{fig:floor}. In the transfer learning phase, the original dense layer with seven classifiers from the pretrained model was replaced with a new linear layer with a softmax output to train the model as a three-classifier suitable for our objective. We divided the collected dataset into a ratio of 8:2 for training and testing, and used the model at epoch 50 for our real demo, achieving the highest test accuracy of 96.3\%.

\begin{figure}[!ht]
    \centering
    \includegraphics[width=0.9\linewidth]{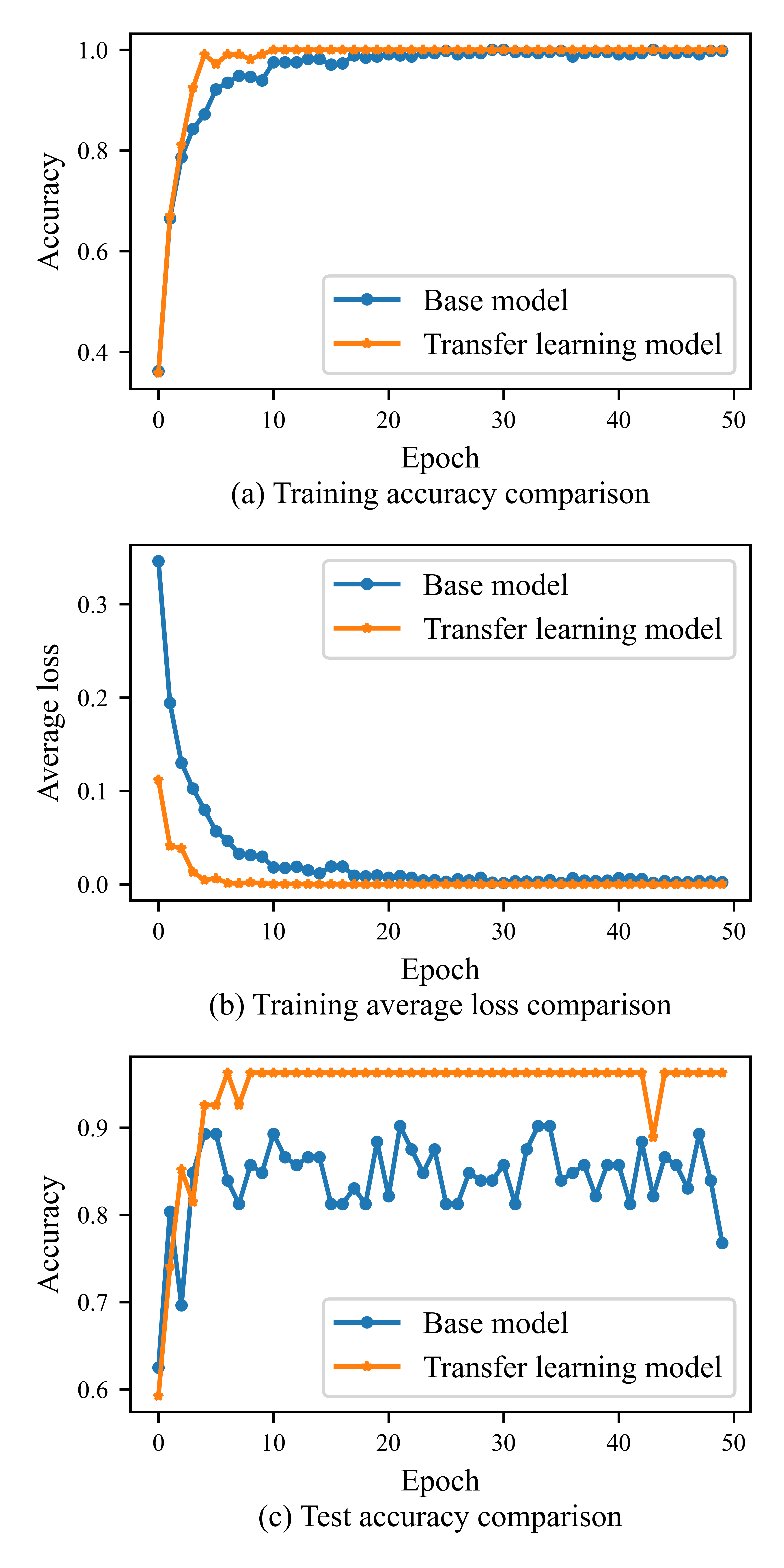}
    \caption{Training accuracy,   average loss and test accuracy comparison of the base model and the transfer learning model.}
    \label{fig:training_comparison}
\end{figure}

\begin{table}[ht]
    \centering
    \begin{minipage}{.5\linewidth}
      \centering
      \caption{Original Dataset Sizes}
      \begin{tabular}{lc}
      \hline
      \textbf{Class} & \textbf{Size} \\
      \hline
      Bed       & 79 \\
      Fall      & 79 \\
      Pick-up    & 80 \\
      Run       & 80 \\
      Sit-down   & 80 \\
      Stand-up   & 79 \\
      Walk      & 80 \\
      \hline 
      \end{tabular} \label{originalDataset}
    \end{minipage}%
    \begin{minipage}{.5\linewidth}
      \centering
      \caption{Transfer Learning Dataset Sizes}
      \begin{tabular}{lc}
      \hline
      \textbf{Class} & \textbf{Size} \\
      \hline
      Fall      & 40 \\
      Normal    & 47 \\
      No-person  & 48 \\
      \hline
      \end{tabular}\label{transferLearningDataset}
    \end{minipage}
\end{table}
\vspace{-10pt}
The experimental results are illustrated in Fig. \ref{fig:training_comparison}, comparing the training accuracy, test accuracy, and average loss of the base model and the transfer learning model.

\begin{table}[ht]
\centering
\caption{Confusion Matrix for Base Model}
\resizebox{\columnwidth}{!}{%
\begin{tabular}{cccccccc}
\hline
 & Bed & Fall & Pick-up & Run & Sit-down & Stand-up & Walk \\
\hline
Bed & 0.82 & 0.06 & 0.00 & 0.00 & 0.06 & 0.06 & 0.00 \\
\hline
Fall & 0.05 & 0.95 & 0.00 & 0.00 & 0.00 & 0.00 & 0.00 \\
\hline
Pick-up & 0.00 & 0.00 & 0.94 & 0.00 & 0.06 & 0.00 & 0.00 \\
\hline
Run & 0.00 & 0.00 & 0.00 & 0.83 & 0.00 & 0.17 & 0.00 \\
\hline
Sit-down & 0.00 & 0.00 & 0.00 & 0.00 & 1.00 & 0.00 & 0.00 \\
\hline
Stand-up & 0.00 & 0.00 & 0.00 & 0.00 & 0.05 & 0.95 & 0.00 \\
\hline
Walk & 0.00 & 0.08 & 0.00 & 0.08 & 0.00 & 0.08 & 0.77 \\
\hline
\end{tabular}
}
\label{BaseModelConfuse}
\end{table}

\begin{table}[ht]
\centering
\caption{Confusion Matrix for Transfer Learning Model}
\begin{tabular}{cccc}
\hline
 & Fall & Walking & No-person/static \\
\hline
Fall & 0.90 & 0.10 & 0.00 \\
\hline
Walking & 0.00 & 1.00 & 0.00 \\
\hline
No-person/static & 0.00 & 0.00 & 1.00 \\
\hline
\end{tabular}\label{TransferLearningConfuse}
\end{table}

The transfer learning model exhibits  improvements over the base model in terms of training speed, average loss and test accuracy. Fig. \ref{fig:training_comparison} shows that the transfer learning model achieves higher accuracy faster and maintains a lower loss throughout the training process. The confusion matrices over the test sets can be found in Tables \ref{BaseModelConfuse} and \ref{TransferLearningConfuse}.

To validate the response effectiveness of the mobile companion robot, we conducted a fall-rescue experiment. In this experiment, a participant performed normal activities and simulated a fall in Room 424. Meanwhile, a robot in Room 422 was carrying out its regular tasks, patrolling between two points. Upon receiving a fall alarm from the server, the robot proceeded to Room 424, removing any obstacles in its path (unlocking, pushing, and traversing a closed door), and carried out the rescue operation. The time from fall detection to the robot's arrival at the rescue location was within three minutes. The experiment included eight trials, of which seven successfully detected and responded to the fall, while one was misidentified as walking, resulting in an overall success rate of 87\%.

Overall, the integrated system demonstrates the potential of utilizing Wi-Fi and mobile robots for fall detection and response, validating the effectiveness of the proposed approach.

\section{Conclusion}
The integration of Wi-Fi sensing with robotic assistance offers a promising solution for fall detection and response. The proposed system provides non-intrusive monitoring, timely detection, and assistance, reducing the risk of long-term injuries. Future work will focus on expanding the system's capabilities and improving the accuracy of detection.
\vspace{-6pt}

\section*{Appendix}

The test demo video is available at 
\href{https://youtu.be/kO990mkyVdc}{[link]}. The collected dataset for transfer learning and the trained model can be accessed at \href{https://drive.google.com/drive/folders/1J5R6a5ybqtAvo4QzWDOLFH36gMliA0V2?usp=sharing}{[link]}.

 \vspace{-6pt}
\section*{Acknowledgment}
This work is partially supported by Shenzhen Key Laboratory of Robotics Perception and Intelligence (ZDSYS20200810171800001), Shenzhen Science and Technology Program under Grant RCBS 20221008093305007, 20231115141459001, Young Elite Scientists Sponsorship Program by CAST under Grant 2023QNRC001, High level of special funds (G03034K003) from Southern University of Science and Technology, Shenzhen, China.
\vspace{-6pt}
\bibliographystyle{IEEEtran}
\bibliography{references}

\end{document}